\documentclass[11pt,a4paper]{article}
\usepackage[hyperref]{acl2021}
\usepackage{times}
\usepackage{latexsym}
\usepackage{todonotes}

\usepackage{covington}
\usepackage{subcaption}
\usepackage{graphicx}
\usepackage{multirow}
\usepackage{adjustbox}
\usepackage{booktabs}

\usepackage{microtype}

\aclfinalcopy 
\def\aclpaperid{2065} 

\title{Lexical Semantic Change Discovery}

\author{Sinan Kurtyigit$^{\spadesuit}$~~~Maike Park$^{\heartsuit}$~~~Dominik Schlechtweg$^{\spadesuit}$\\
\textbf{Jonas Kuhn$^{\spadesuit}$~~~Sabine Schulte im Walde}$^{\spadesuit}$\\
\\
$^{\spadesuit}$Institute for Natural Language Processing, University of Stuttgart\\
$^{\heartsuit}$Leibniz Institute for the German Language, Mannheim\\
\small 
{\tt sinan.kurtyigit@gmail.com},
\small
{\tt park@ids-mannheim.de}\\
\small
{\tt \{schlecdk,jonas.kuhn,schulte\}@ims.uni-stuttgart.de}}

\date{}

\begin{document}
\maketitle
\begin{abstract}
While there is a large amount of research in the field of Lexical Semantic Change Detection, only few approaches go beyond a standard benchmark evaluation of existing models. In this paper, we propose a shift of focus from change detection to change discovery, i.e., discovering novel word senses over time from the full corpus vocabulary. By heavily fine-tuning a type-based and a token-based approach on recently published German data, we demonstrate that both models can successfully be applied to discover new words undergoing meaning change. Furthermore, we provide an almost fully automated framework for both evaluation and discovery.
\end{abstract}

\section{Introduction}
There has been considerable progress in Lexical Semantic Change Detection (LSCD) in recent years \citep{kutuzov-etal-2018-diachronic,2018arXiv181106278T,hengchen2021challenges}, with milestones such as the first approaches using neural language models \citep{Kim14,Kulkarni14}, the introduction of Orthogonal Procrustes alignment \citep{Kulkarni14,Hamilton:2016}, detecting sources of noise \citep{dubossarsky2017,Dubossarskyetal19}, the formulation of continuous models \citep{Frermann:2016,RosenfeldE18,tsakalidis-liakata-2020-sequential}, the first uses of contextualized embeddings \citep{Hu19,giulianelli-etal-2020-analysing}, the development of solid annotation and evaluation frameworks \citep{Schlechtwegetal18,Schlechtwegetal19,Shoemark2019} and shared tasks \citep{diacrita_evalita2020,schlechtweg-etal-2020-semeval}.

However, only a very limited amount of work applies the methods to \textbf{discover novel instances of semantic change} and to evaluate the usefulness of such discovered senses for external fields. That is, the majority of research focuses on the introduction of novel LSCD models, and on analyzing and evaluating existing models. Up to now, these preferences for development and analysis vs. application represented a well-motivated choice, because the quality of state-of-the-art models had not been established yet, and because no tuning and testing data were available. But with recent advances in evaluation \citep{diacrita_evalita2020,schlechtweg-etal-2020-semeval,rushifteval2021}, the field now owns standard corpora and tuning data for different languages. Furthermore, we have gained experience regarding the interaction of model parameters and modelling task (such as binary vs. graded semantic change). This enables the field to more confidently apply models to discover previously unknown semantic changes. Such discoveries may be useful in a range of fields \citep{hengchen2019nation,jatowt2021applications}, among which historical semantics and lexicography represent obvious choices \citep{Ljubesic2020deep}.

In this paper, we tune the most successful models from SemEval-2020 Task 1 \citep{schlechtweg-etal-2020-semeval} on the German task data set in order to obtain high-quality discovery predictions for novel semantic changes.
We validate the model predictions in a standardized human annotation procedure and visualize the annotations in an intuitive way supporting further analysis of the semantic structure relating word usages. In this way, we automatically detect previously described semantic changes and at the same time discover novel instances of semantic change which had not been indexed in standard historical dictionaries before. Our approach is largely automated, by relying on unsupervized language models and a publicly available annotation system requiring only a small set of judgments from annotators. We further evaluate the usability of the approach from a lexicographer's viewpoint and show how intuitive visualizations of human-annotated data can benefit dictionary makers.

\section{Related Work}
State-of-the-art semantic change detection models are Vector Space Models (VSMs) \citep{schlechtweg-etal-2020-semeval}. These can be divided into type-based (static) \citep{Turney:2010} and token-based (contextualized) \citep{Schutze1998} approaches. For our study, we use both a static and a contextualized model. As mentioned above, previous work mostly focuses on creating data sets or developing, evaluating and analyzing models. A common approach for evaluation is to annotate target words selected from dictionaries in specific corpora \citep{Tahmasebi17,Schlechtwegetal18,Perrone19,diacrita_evalita2020,rodina2020rusemshift,schlechtweg-etal-2020-semeval}. Contrary to this, our goal is to find `undiscovered' changing words and validate the predictions of our models by human annotators. Few studies focus on this task. \citet{Kim14}, \citet{Hamilton:2016}, \citet{basile2016diachronic}, \citet{basile2018exploiting}, \citet{TakamuraEtal2017} and \citet{tsakalidis-etal-2019-mining} evaluate their approaches by validating the top ranked words through author intuitions or known historical data. The only approaches applying a systematic annotation process are \citet{Gulordava11} and \citet{Cook13p49}. \citeauthor{Gulordava11} ask human annotators to rate 100 randomly sampled words on a 4-point scale from 0 (no change) to 3 (changed significantly), however without relating this to a data set. \citeauthor{Cook13p49} work closely with a professional lexicographer to inspect 20 lemmas predicted by their models plus 10 randomly selected ones. \citeauthor{Gulordava11} and \citeauthor{Cook13p49} evaluate their predictions on the (macro) lemma level. We, however, annotate our predictions on the (micro) usage level, enabling us to better control the criteria for annotation and their inter-subjectivity. In this way, we are also able to build clusters of usages with the same sense and to visualise the annotated data in an intuitive way. The annotation process is designed to not only improve the quality of the annotations, but also lessen the burden on the annotators. We additionally seek the opinion of a professional lexicographer to assess the usefulness of the predictions outside the field of LSCD. 

In contrast to previous work, we obtain model predictions by fine-tuning static and contextualized embeddings on high-quality data sets \citep{schlechtweg-etal-2020-semeval} that were not available before. We provide a highly automated general framework for evaluating models and predicting changing words on all kinds of corpora.

\vspace{+1mm}
\section{Data}
We use the German data set provided by the SemEval-2020 shared task \citep{schlechtweg-etal-2020-semeval,Schlechtweg2021dwug}. The data set contains a diachronic corpus pair for two time periods to be compared, a set of carefully selected target words as well as binary and graded gold data for semantic change evaluation and fine-tuning purposes. 

\paragraph{Corpora}
The DTA corpus \citep{dta2017} and a combination of the BZ \citep{BZ2018} and ND \citep{ND2018} corpora are used. DTA contains texts from different genres spanning the 16th--20th centuries. BZ and ND are newspaper corpora jointly spanning 1945--1993. \citet{schlechtweg-etal-2020-semeval} extract two time specific corpora $C_1$ (DTA, 1800--1899) and $C_2$ (BZ+ND 1946--1990) and provide raw and lemmatized versions. 

\paragraph{Target Words}
A list of 48 target words, consisting of 32 nouns, 14 verbs and 2 adjectives is provided. These are controlled for word frequency to minimize model biases that may lead to artificially high performance \citep{dubossarsky2017, schlechtweg-walde-2020}.

\vspace{+1mm}
\section{Models}
Type-based models generate a single vector for each word from a pre-defined vocabulary. In contrast, token-based models generate one vector for each usage of a word. While the former do not take into account that most words have multiple senses, the latter are able to capture this particular aspect and are thus presumably more suited for the task of LSCD \citep{Martinc2020evolution}. Even though contextualized approaches have indeed significantly outperformed static approaches in several NLP tasks over the past years \citep{ethayarajh2019contextual}, the field of LSCD is still dominated by type-based models \citep{schlechtweg-etal-2020-semeval}.  \citet{kutuzov-giulianelli-2020-uiouva} yet show that the performance of token-based models (especially ELMo) can be increased by fine-tuning on the target corpora.
\citet{laicher-etal-2020-volente,Laicher2021explaining} drastically improve the performance of BERT by reducing the influence of target word morphology. In this paper, we compare both families of approaches for change discovery.

\subsection{Type-based approach}
Most type-based approaches in LSCD combine three sub-systems: (i) creating semantic word representations, (ii) aligning them across corpora, and (iii) measuring differences between the aligned representations \citep{Schlechtwegetal19}. Motivated by its wide usage and high performance among participants in SemEval-2020 \citep{schlechtweg-etal-2020-semeval} and DIACR-Ita \citep{diacrita_evalita2020}, we use the Skip-gram with Negative Sampling model \citep[SGNS,][]{Mikolov13a, Mikolov13b} to create static word embeddings. SGNS is a shallow neural language model trained on pairs of word co-occurrences extracted from a corpus with a symmetric window. The optimized parameters can be interpreted as a semantic vector space that contains the word vectors for all words in the vocabulary. In our case, we obtain two separately trained vector spaces, one for each subcorpus ($C_1$ and $C_2$). Following standard practice, both spaces are length-normalized, mean-centered \citep{artetxe2016,Schlechtwegetal19} and then aligned by applying Orthogonal Procrustes (OP), because columns from different vector spaces may not correspond to the same coordinate axes \citep{Hamilton:2016}. The change between two time-specific embeddings is measured by calculating their Cosine Distance (CD) \citep{SaltonMcGill1983}. The strength of SGNS+OP+CD has been shown in two recent shared tasks with this sub-system combination ranking among the best submissions \citep{arefyev-zhikov-2020-cmce, kaiser-etal-2020-roots, pomsl-lyapin-2020-circe, prazak-etal-2020-CCA}.

\subsection{Token-based approach}
\label{sec:token-based}
Bidirectional Encoder Representations from Transformers \citep[BERT,][]{devlin-etal-2019-bert} is a transformer-based neural language model designed to find contextualized representations for text by analyzing left and right contexts. The base version processes text in 12 different layers. In each layer, a contextualized token vector representation is created for every word. A layer, or a combination of multiple layers (we use the average), then serves as a representation for a token. For every target word we extract usages (i.e., sentences in which the word appears) by randomly sub-sampling up to 100 sentences from both subcorpora $C_1$ and $C_2$.\footnote{We sub-sample as some words appear in 10,000 or more sentences.} These are then fed into BERT to create contextualized embeddings,  resulting in two sets of up to 100 contextualized vectors for both time periods. To measure the change between these sets we use two different approaches: (i) We calculate the Average Pairwise Distance (APD). The idea is to randomly pick a number of vectors from both sets and measure their mutual distances \citep{Schlechtwegetal18,kutuzov-giulianelli-2020-uiouva}. The change score corresponds to the mean average distance of all comparisons. (ii) We average both vector sets and measure the Cosine Distance (COS) between the two resulting mean vectors \citep{kutuzov-giulianelli-2020-uiouva}.

\section{Discovery}
\label{sec:tuning}

SemEval-2020 Task 1 consists of two subtasks: (i)~binary classification: for a set of target words, decide whether (or not) the words lost or gained sense(s) between $C_1$ and $C_2$, and (ii)~graded ranking: rank a set of target words according to their degree of LSC between $C_1$ and $C_2$. These require to detect semantic change in a small pre-selected set of target words. Instead, we are interested in the discovery of changing words from the full vocabulary of the corpus. We define the task of \textbf{lexical semantic change discovery} as follows.
\begin{itemize}
  \item[] Given a diachronic corpus pair $C_1$ and $C_2$, decide for the intersection of their vocabularies which words lost or gained sense(s) between $C_1$ and $C_2$.
\end{itemize}
This task can also be seen as a special case of 
SemEval's Subtask 1 where the target words equal the intersection of the corpus vocabularies. Note, however, that discovery introduces additional difficulties for models,  e.g. because a large number of predictions is required and the target words are not preselected, balanced or cleaned. Yet, discovery is an important task, with applications such as lexicography where dictionary makers aim to cover the full vocabulary of a language.

\subsection{Approach}

We start the discovery process by generating optimized graded value predictions using high-performing parameter configurations following previous work and fine-tuning. Afterwards, we infer binary scores with a thresholding technique (see below). We then tune the threshold to find the best-performing type- and token-based approach for binary classification. These are used to generate two sets of predictions.\footnote{Find the code used for each step of the prediction process at \url{https://github.com/seinan9/LSCDiscovery}.}

\paragraph{Evaluation metrics}
We evaluate the graded rankings in Subtask 2 by computing Spearman's rank-order correlation coefficient $\rho$. For the binary classification subtask we compute precision, recall and F$_{0.5}$. The latter puts a stronger focus on precision than recall because our human evaluation cannot be automated, so we decided to weigh quality (precision) higher than quantity (recall).

\paragraph{Parameter tuning}
Solving Subtask 2 is straightforward, since both the type-based and token-based approaches output distances between representations for $C_1$ and $C_2$ for every target word. Like many approaches in SemEval-2020 Task 1 and DIACR-Ita we use thresholding to binarise these values. The idea is to define a threshold parameter, where all ranked words with a distance greater or equal to this threshold are labeled as changing words. 

For cases where no tuning data is available, \citet{kaiser-etal-2020-roots} propose to choose the threshold according to the population of CDs of all words in the corpus. \citeauthor{kaiser-etal-2020-roots} set the threshold to $\mu + \sigma$, where $\mu$ is the mean and $\sigma$ is the standard deviation of the population. We slightly modify this approach by changing the threshold to $\mu + t * \sigma$. In this way, we introduce an additional parameter $t$, which we tune on the SemEval-2020 test data. We test different values ranging from $-2$ to $2$ in steps of $0.1$.

\begin{table}[t]
\centering
\tabcolsep=0.11cm
\begin{tabular}{ll}
\multirow{4}{*}{$\Bigg\uparrow$} &4: Identical\\
 &3: Closely Related\\
 &2: Distantly Related\\
 &1: Unrelated\\
\end{tabular}
\caption{DURel relatedness scale \citep{Schlechtwegetal18}.}\label{tab:scales}
\vspace{-8ex} 
\end{table}

\paragraph{Population}
Since SGNS generates type-based vectors for every word in the vocabulary, measuring the distances for the full vocabulary comes with low additional computational effort. Unfortunately, this is much more difficult for BERT. Creating up to 100 vectors for every word in the vocabulary drastically increases the computational burden. We choose a population of 500 words for our work allowing us to test multiple parameter configurations.\footnote{In a practical setting where predictions have to be generated only once, a much larger number may be chosen. Also, possibilities to scale up BERT performance can be applied \citep{Montariol2021scalable}.} We sample words from different frequency areas to have predictions not only for low-frequency words. For this, we first compute the frequency range (highest frequency -- lowest frequency) of the vocabulary. This range is then split into 5 areas of equal frequency width. Random samples from these areas are taken based on how many words they contain. For example: if the lowest frequency area contains 50\% of all words from the vocabulary, then $0.5 * 500 = 250$ random samples are taken from this area. The SemEval-2020 target words are excluded from this sampling process. The resulting population is used to create predictions for both models. 

\paragraph{Filtering}
The predictions contain proper names, foreign language and lemmatization errors, which we aim to filter out, as such cases are usually not considered as semantic changes. We only allow nouns, verbs and adjectives to pass. Words where over 10\% of the usages are either non-German or contain more than 25\% punctuation are filtered out as well.

\begin{figure*}[t]
    \begin{subfigure}{0.33\textwidth}
\frame {        \includegraphics[width=\linewidth]{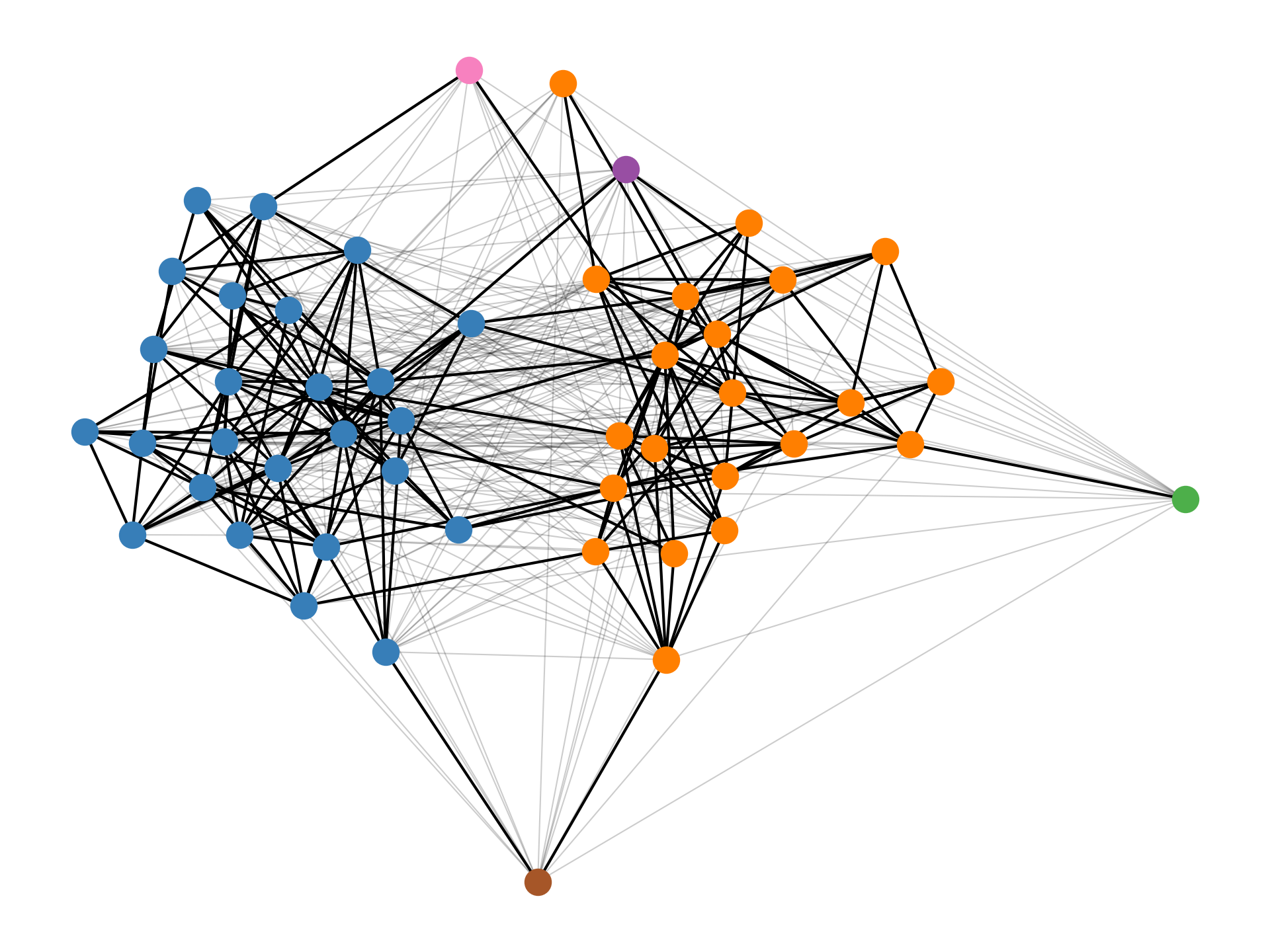}}
        \caption*{full}
    \end{subfigure}
    \begin{subfigure}{0.33\textwidth}
\frame{        \includegraphics[width=\linewidth]{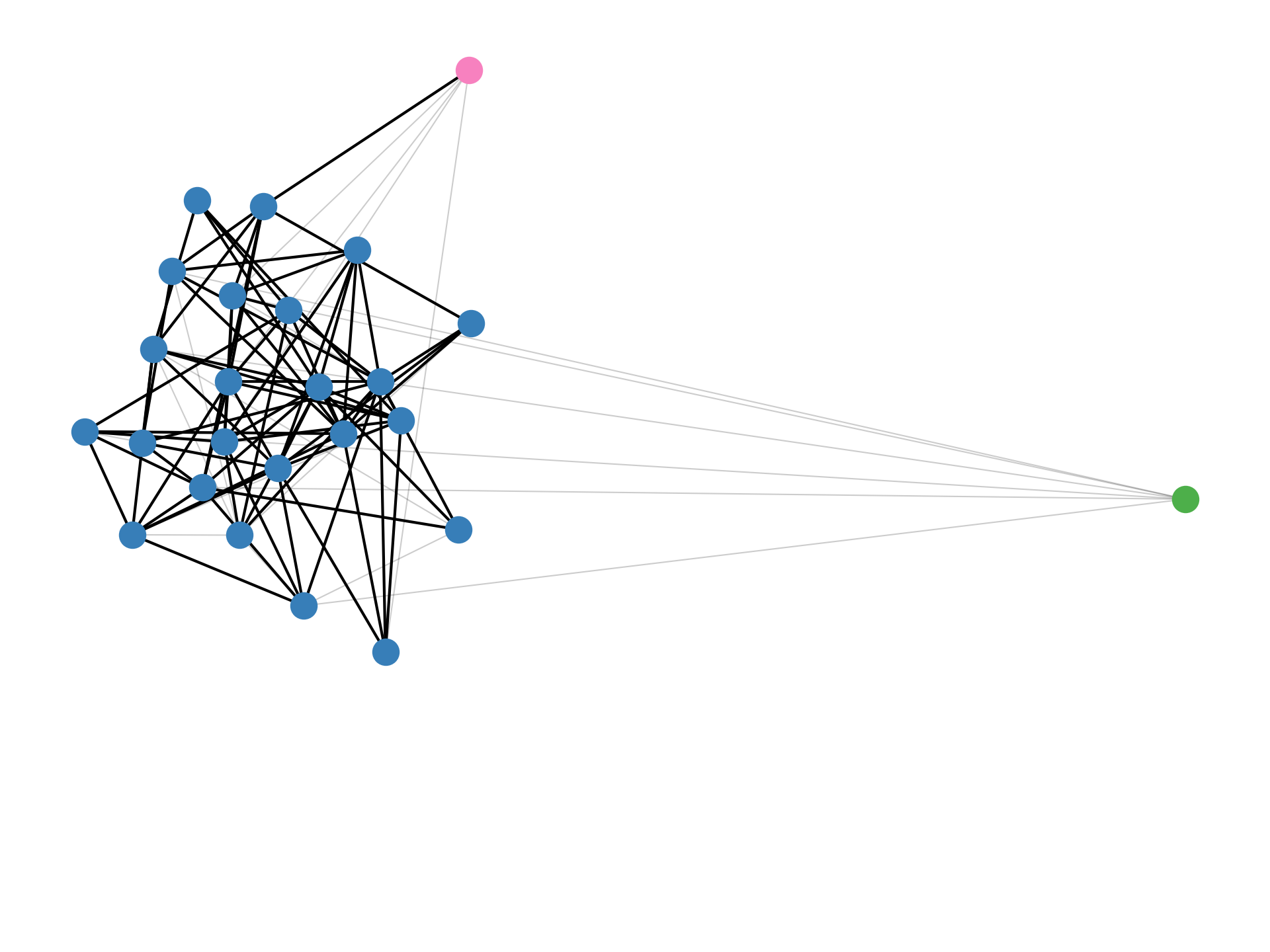}}
        \caption*{$C_1$}
    \end{subfigure}
    \begin{subfigure}{0.33\textwidth}
\frame{        \includegraphics[width=\linewidth]{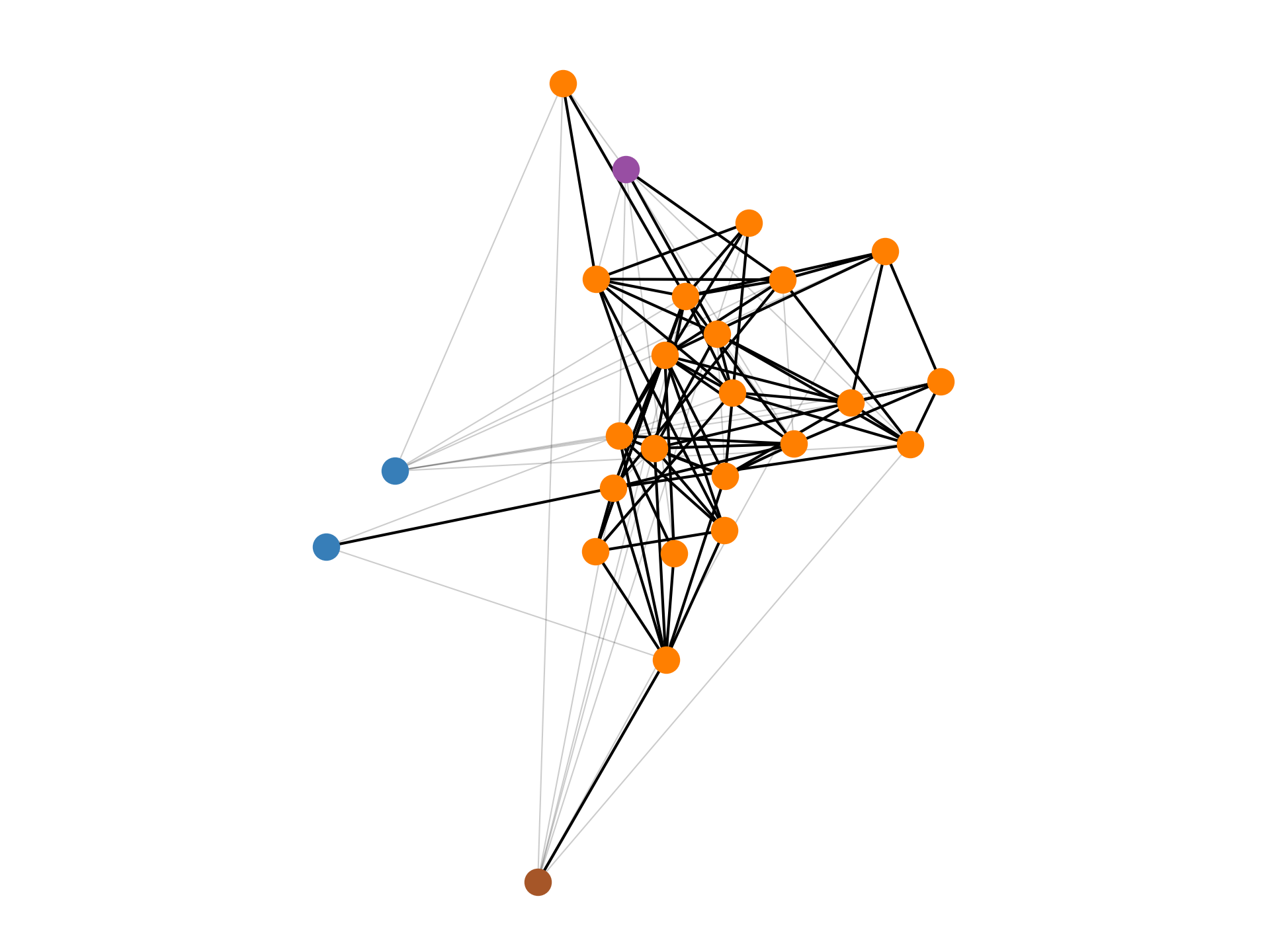}}
        \caption*{$C_2$}
    \end{subfigure}
    \caption{Word Usage Graph of German \textit{Aufkommen} (left), subgraphs for first time period $C_1$ (middle) and for second time period $C_2$ (right). \textbf{black}/\textcolor{gray}{gray} lines indicate \textbf{high}/\textcolor{gray}{low} edge weights.}\label{fig:aufkommen}
\end{figure*}

\section{Annotation}
\label{sec:anno}

The model predictions are validated by human annotation. For this, we apply the SemEval-2020 Task 1 procedure, as described in \citet{schlechtweg-etal-2020-semeval}. Annotators are asked to judge the semantic relatedness of pairs of word usages, such as the two usages of \textit{Aufkommen} in (\ref{ex:1}) and (\ref{ex:2}), on the scale in Table \ref{tab:scales}. 
\begin{example}\label{ex:1}
Es ist richtig, dass mit dem \textbf{Aufkommen} der Manufaktur im Unterschied zum Handwerk sich Spuren der Kinderexploitation zeigen.\\
`\em It is true that with the \textbf{emergence} of the manufactory, in contrast to the handicraft, traces of child labor are showing.'
\end{example}%
\begin{example}\label{ex:2}
Sie wissen, daß wir für das Vieh mehr Futter aus eigenem \textbf{Aufkommen} brauchen.\\
`\em They know that we need more feed from our own \textbf{production} for the cattle.'
\end{example}
The annotated data of a word is represented in a Word Usage Graph (WUG), where vertices represent word usages, and weights on edges represent the (median) semantic relatedness judgment of a pair of usages such as (\ref{ex:1}) and (\ref{ex:2}). The final WUGs are clustered with a variation of correlation clustering \citep{Bansal04,schlechtweg-etal-2020-semeval} (see Figure \ref{fig:aufkommen}, left) and split into two subgraphs representing nodes from subcorpora $C_1$ and $C_2$, respectively (middle and right). Clusters are then interpreted as word senses and changes in clusters over time as lexical semantic change.

In contrast to \citeauthor{schlechtweg-etal-2020-semeval} we use the openly available DURel interface for annotation and visualization.\footnote{\url{https://www.ims.uni-stuttgart.de/data/durel-tool}.} This also implies a change in sampling procedure, as the system currently implements only random sampling of use pairs (without SemEval-style optimization). For each target word we sample $|U_1|=|U_2|=25$ usages (sentences) per subcorpus ($C_1$, $C_2$) and upload these to the DURel system, which presents use pairs to annotators in randomized order. We recruit eight German native speakers with university level education as annotators. Five have a background in linguistics, two in German studies, and one has an additional professional background in lexicography. Similar to \citeauthor{schlechtweg-etal-2020-semeval}, we ensure the robustness of the obtained clusterings by continuing the annotation of a target word until all multi-clusters (clusters with more than one usage) in its WUG are connected by at least one judgment. We finally label a target word as changed (binary) if it gained or lost a cluster over time. For instance, \textit{Aufkommen} in Figure \ref{fig:aufkommen} is labeled as change as it gains the orange cluster from $C_1$ to $C_2$. Following \citet{schlechtweg-etal-2020-semeval} we use $k$ and $n$ as lower frequency thresholds to avoid that small random fluctuations in sense frequencies caused by sampling variability or annotation error be misclassified as change. As proposed in \citet{Schlechtweg2021wugs} for comparability across sample sizes we set $k = 1 \leq 0.01*|U_i| \leq 3$ and $n = 3 \leq 0.1*|U_i| \leq 5$, where $|U_i|$ is the number of usages from the respective time period (after removing incomprehensible usages from the graphs). This results in $k=1$ and $n=3$ for all target words.

Find an overview over the final set of WUGs in Table \ref{tab:data}. We reach a comparably high inter-annotator agreement (Krippendorf's $\alpha=.58$).\footnote{We provide WUGs as Python NetworkX graphs, descriptive statistics, inferred clusterings, change values and interactive visualizations for all target words and the respective code at \url{https://www.ims.uni-stuttgart.de/data/wugs}.}

\begin{table*}[t]
\centering
\tabcolsep=.09cm
\begin{tabular}{ c | c c c c c c c c c c c c}            
\toprule                           
\textbf{Data set}  &  $\mathbf{n}$  &  \textbf{N/V/A } &  $\mathbf{|U|}$  & \textbf{AN} & \textbf{ JUD } &  \textbf{AV } &  \textbf{SPR } &  \textbf{KRI } &  \textbf{UNC } & \textbf{ LOSS } & \textbf{LSC$_B$}& \textbf{LSC$_G$} \\
\midrule                           
SemEval & 48 &  32/14/2  & 178 & 8 &  37k  & 2 & .59 & .53 & 0 & .12 & .35 & .31 \\
Predictions & 75 & 39/16/20 & 49 & 8 & 24k & 1 & .64 & .58 & 0 & .26 & .48 & .40 \\
\bottomrule                           
\end{tabular}
\caption{Overview target words. $n$ = no. of target words, N/V/A = no. of nouns/verbs/adjectives, $|U|$ = avg. no. of usages per word, AN = no. of annotators, JUD = total no. of judged usage pairs, AV = avg. no. of judgments per usage pair, SPR = weighted mean of pairwise Spearman, KRI = Krippendorff's $\alpha$, UNC = avg. no. of uncompared multi-cluster combinations, LOSS = avg. of normalized clustering loss * 10, LSC$_{B/G}$ = mean binary/graded change score.}
\label{tab:data}
\end{table*}

\section{Results}
We now describe the results of the tuning and discovery procedures. 

\subsection{Tuning}
SGNS is commonly used \citep{schlechtweg-etal-2020-semeval} and also highly optimized \citep{kaiser-etal-2020-IMS, kaiser-etal-2020-roots, Kaiser2021effects}, so it is difficult to further increase the performance. We thus rely on the work of \citet{kaiser-etal-2020-IMS} and test their parameter configurations on the German SemEval-2020 data set.\footnote{All configurations use $w=10$, $d=300$, $e=5$ and a minimum frequency count of $39$.} We obtain three slightly different parameter configurations (see Table \ref{tab:param} for more details), yielding competitive $\rho=.690$, $\rho=.710$ and $\rho=.710$, respectively.  

In order to improve the performance of BERT, we test different layer combinations, pre-processings and semantic change measures. Following \citet{laicher-etal-2020-volente,Laicher2021explaining}, we are able to drastically increase the performance of BERT on the German SemEval-2020 data. In a pre-processing step, we replace the target word in every usage by its lemma. In combination with layer 12+1, both APD and COS perform competitively well on Subtask 2 ($\rho=.690$ and $\rho=.738$). 

After applying thresholding as described in Section \ref{sec:tuning} we obtain F$_{0.5}$-scores for a large range of thresholds. SGNS achieves peak F$_{0.5}$-scores of $.692$, $.738$ and $.685$, respectively (see Table \ref{tab:param}). Interestingly, the optimal threshold is at $t=1.0$ in all three cases. This corresponds to the threshold used in \citet{kaiser-etal-2020-roots}. While the peak F$_{0.5}$ of BERT+APD is marginally worse ($.598$ at $t=-0.2$), BERT+COS is able to outperform the best SGNS configuration with a peak of $.741$ at $t=0.1$.   

In order to obtain an estimate on the sampling variability that is caused by sampling only up to 100 usages per word for BERT+APD and BERT+COS (see Section \ref{sec:token-based}), we repeat the whole procedure 9 times and estimate mean and standard deviation of performance on the tuning data. In the beginning of every run the usages are randomly sampled from the corpora. We observe a mean $\rho$ of $.657$ for BERT+APD and $.743$ for BERT+COS with a standard deviation of $.015$ and $.012$, respectively, as well as a mean F$_{0.5}$ of $.576$ for BERT+APD and $.684$ for BERT+COS with a standard deviation of $.013$ and $.038$, respectively. This shows that the variability caused by sub-sampling word usages is negligible.

\begin{table*}[t]
\centering
\begin{adjustbox}{width=0.8\textwidth}
\begin{tabular}{ c | c | c | c c c c | c c c c }
\toprule
& \multirow{2}{*}{parameters} & \multirow{2}{*}{$t$} & \multicolumn{4}{c|}{\textbf{tuning}} & \multicolumn{4}{c}{\textbf{predictions}} \\
& & & $\rho$ & F$_{0.5}$ & P & R & $\rho$ & F$_{0.5}$ & P & R \\
\midrule
\multirow{3}{*}{\rotatebox[origin=c]{90}{\small\textbf{SGNS}}} & \small$k=1$, \small$s=.005$ & $1.0$ & $.690$ & $.692$ & $.750$ & $.529$ \\
& \small$\mathbf{k=5, s=.001}$ & $1.0$ & $.710$ & $\mathbf{.738}$ & $.818$ & $.529$ & $.295$ & $\mathbf{.714}$ & $.667$ & $1.0$\\
& \small$k=5$, $s=$ None & $1.0$ & $.710$ & $.685$ & $.714$ & $.588$ \\
\midrule
\multirow{2}{*}{\rotatebox[origin=c]{90}{\small\textbf{BERT}}} & \small{APD} & $-0.2$ & $.673$ & $.598$ & $.560$ & $.824$ \\
& \small\textbf{COS} & $0.1$ & $.738$ & $\mathbf{.741}$ & $.706$ & $.788$ & $.482$ & $.620$ &$.567$ & $1.0$\\
\midrule
\multirow{1}{*}{\rotatebox[origin=c]{90}{\small\textbf{BL}}} & \small{random sampling} & & & & & & & $.349$ & $.300$ & $1.0$ \\
\bottomrule
\end{tabular}
\end{adjustbox}
\caption{Performance (Spearman $\rho$, F$_{0.5}$-measure, precission P and recall R) of different approaches on tuning data (SemEval targets) and performance of best type- and token-based approach on respective predictions with optimal tuning threshold $t$, as well as the performance of a randomly sampled baseline.}
\label{tab:param}
  \end{table*}

\subsection{Discovery}
We use the top-performing configurations (see Table \ref{tab:param}) to generate two sets of large-scale predictions. While we use the lemmatized corpora for SGNS, in BERT's case we choose the raw corpora with lemmatized target words instead. The latter choice is motivated by the previously described performance increases. After the filtering as described in Section \ref{sec:anno}, we obtain 27 and 75 words labeled as changing, respectively. We further sample 30 targets from the second set of predictions to obtain a feasible number for annotation. We call the first set SGNS targets and the second one BERT targets, with an overlap of 7 targets. Additionally, we randomly sample 30 words from the population (with an overlap of 5 with the SGNS and BERT targets) in order to have an indication of what the change distribution underlying the corpora is. We call these baseline (BL) targets. This baseline will help us to put the results of the predictions in context and to find out whether the predictions of the two models can be explained by pure randomness. Following the annotation process, binary gold data is generated for all three target sets, in order to validate the quality of the predictions.

The evaluation of the predictions is presented in Table \ref{tab:param}. We achieve a F$_{0.5}$-score of $.714$ for SGNS and $.620$ for BERT. Out of the 27 words predicted by the SGNS model, 18 (67 \%) were actually labeled as changing words by the human annotators. In comparison, only 17 out of the 30 (57 \%) BERT predictions were annotated as such. The performance of SGNS for prediction (SGNS targets) is even higher than on the tuning data (SemEval targets). In contrast, BERT's performance for prediction drops strongly in comparison to the performance on the tuning data ($.741$ vs. $.620$). This reproduces previous results and confirms that (off-the-shelf) BERT generalises poorly for LSCD and does not transfer well between data sets \citep{laicher-etal-2020-volente}. If we compare these results to the baseline, we can see that both models perform much better than the random baseline (F$_{0.5}$ of $.349$). Only 10 out of the 30 (30 \%) randomly sampled words are annotated as changing. This indicates, that the performance of SGNS and BERT is likely not a cause of randomness. Both models considerably increase the chance of finding changing words compared to a random model.

\begin{figure*}[t]
    \begin{subfigure}{0.5\textwidth}
        \includegraphics[width=\linewidth]{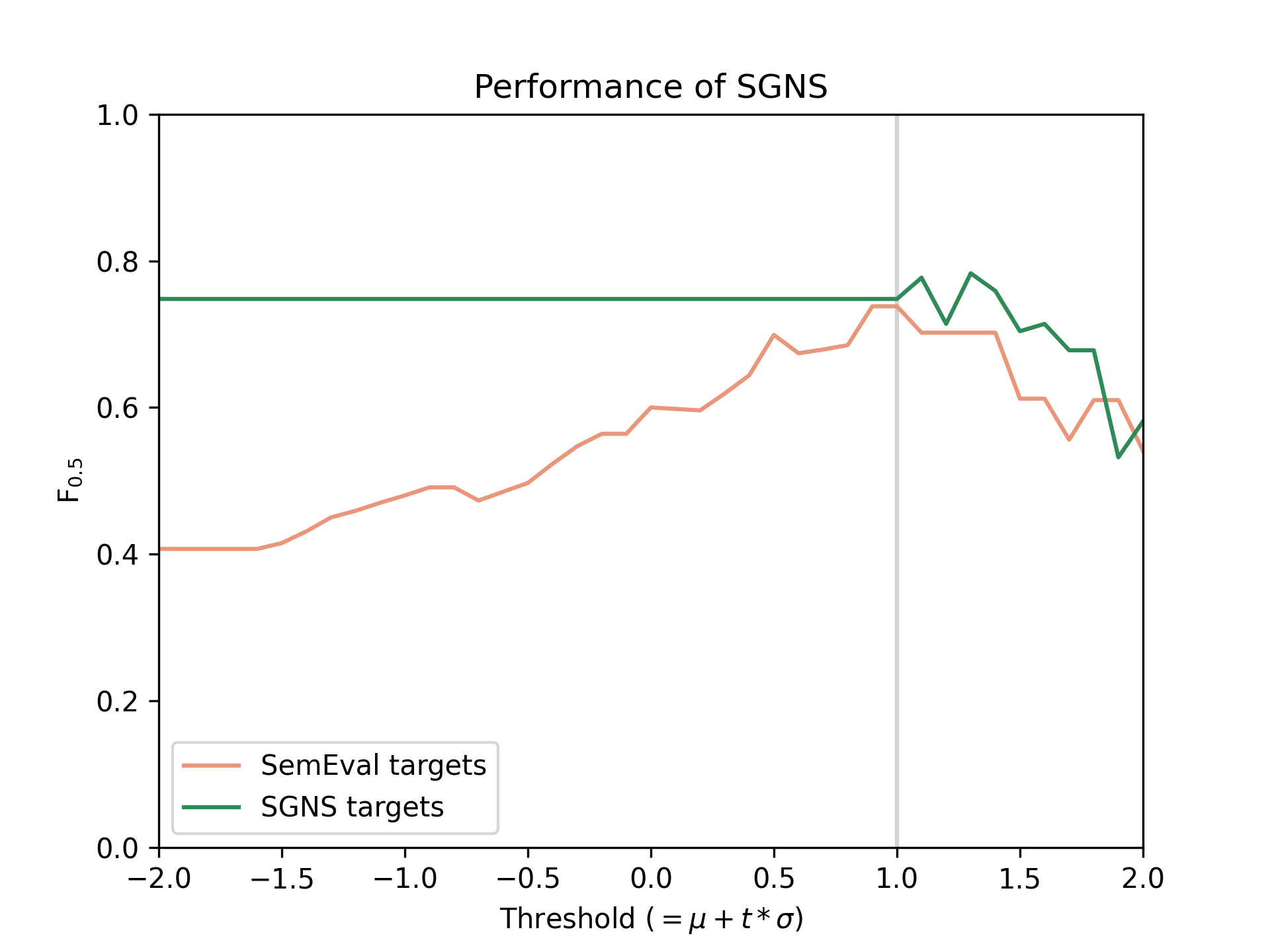}
        \label{fig:sgns_perf}
    \end{subfigure}
    \begin{subfigure}{0.5\textwidth}
        \includegraphics[width=\linewidth]{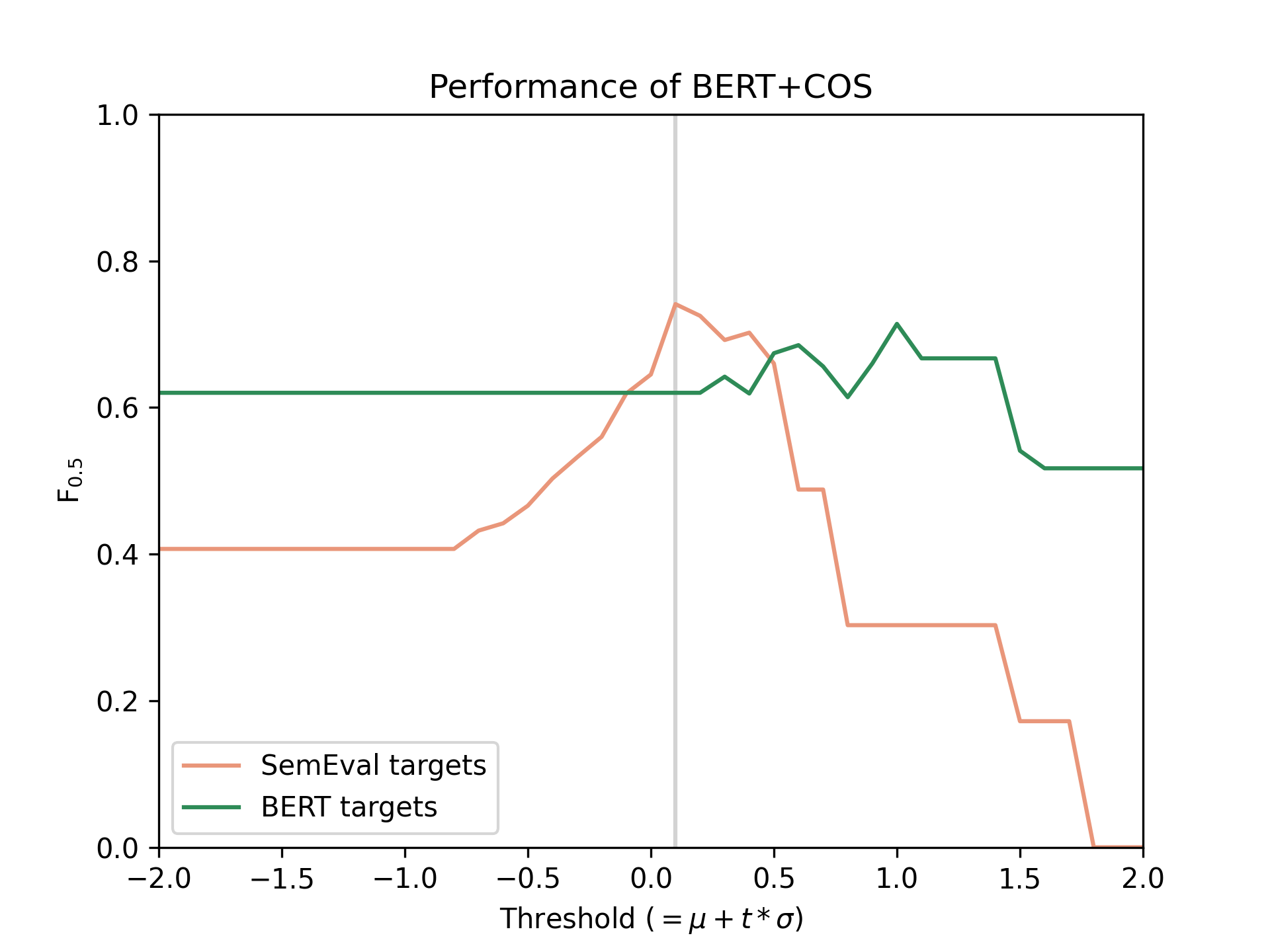}
        \label{fig:cos_perf}
    \end{subfigure}
  \vspace{-20pt}
  \caption{F$_{0.5}$ performance on SemEval targets (orange) and respective predictions (green) across different thresholds. Left: SGNS. Right: BERT+COS. Gray vertical line indicates optimal performance on SemEval targets.}
    \label{fig:label2}
\end{figure*}

Figure \ref{fig:label2} shows the detailed F$_{0.5}$ developments across different thresholds on the SemEval targets and the predicted words. Increasing the threshold on the predicted words improves the F$_{0.5}$ for both the type-based and token-based approach. A new high-score of $.783$ at $t=1.3$ is achievable for SGNS. While BERT's performance also increases to a peak of $.714$ at $t=1.0$, it is still lower than in the tuning phase.

\subsection{Analysis}
For further insights into sources of errors, we take a close look at the false positives, their WUGs and the underlying usages. Most of the wrong predictions can be grouped into one out of two error sources  \citep[cf.][pp.~175--182]{kutuzov2020distributional}.

\paragraph{Context change}
The first category includes words where the context in the usages shifts between time periods, while the meaning stays the same. The WUG of \textit{Angriffswaffe} (`offensive weapon') (see Figure \ref{fig:angriffswaffe} in Appendix \ref{sec:appendix}) shows a single cluster for both $C_1$ and $C_2$. In the first time period \textit{Angriffswaffe} is used to refer to a hand weapon (such as  `sword', `spear'). In the second period, however, the context changes to nuclear weaponry. We can see a clear contextual shift, while the meaning did not change. In this case both models are tricked by the change of context. Further false positives in this category are the SGNS targets \emph{Ächtung} (`ostracism') and \emph{aussterben} (`to die out') and the COS targets \emph{Königreich} (`kingdom') and \emph{Waffenruhe} (`ceasefire'). 

\paragraph{Context variety}
Words that can be used in a large variety of contexts form the second group of false positives. SGNS falsely predicts \emph{neunjährig} as a changing word. We take a closer look at its WUG (see Figure \ref{fig:neunjahrig} in Appendix \ref{sec:appendix}). We observe that there is only one and the same cluster in both time periods, and the meaning of the target does not change, even though a large variety of contexts exists in both $C_1$ and $C_2$. For example: `which bears oats at \textbf{nine years} fertilization', `courageously, a \textbf{nine-year-old} Spaniard did something' and `after nine years of work'. Both models are misguided by this large context variety. Examples include the SGNS targets \emph{neunjährig} (`9-year-old') and \emph{vorjährig} (`of the previous year') and the COS targets \emph{bemerken} (`to notice') and \emph{durchdenken} (`to think through').

\section{Lexicographical evaluation}
We now evaluate the usefulness of the proposed semantic change discovery procedure including the annotation system and WUG visualization from a lexicographer's viewpoint. The advantage of our approach lies in providing lexicographers and dictionary makers the choice to take a look into predictions they consider promising with respect to their research objective (disambiguation of word senses, detection of novel senses, detection of archaisms, describing senses in regard to specific discourses etc.) and the type of dictionary. Visualized predictions for target words may be analyzed in regard to single senses, clusters of senses, the semantic proximity of sense clusters and a stylized representation of frequency. Random sampling of usages also offers the opportunity to judge underrepresented senses in a sample that might be infrequent in a corpus or during a specific period of time (although currently a high number of overall annotations would be required in order to do so). Most importantly, the use of a variable number of human annotators has the potential to ensure a more objective analysis of large amounts of corpus data. 
In order to evaluate the potential of the approach for assisting lexicographers with extending dictionaries, we analyze statistical measures and predictions of the models provided for the two sets of predictions (SGNS, BERT) and compare them to existing dictionary contents. 

We consider overall inter-annotator agreement ($\alpha>=.5$) and annotated binary change label to select 21 target words for lexicographical analysis. In this way, we exclude unclear cases and non-changing words. The target words are analyzed by inspecting cluster visualizations of WUGs (such as in Figure \ref{fig:aufkommen}) and comparing them to entries in general and specialized dictionaries in order to determine: 
\begin{itemize}
    \item whether a candidate novel sense is already included in one of the reference dictionaries,
    \item whether a candidate novel sense is included in one of the two reference dictionaries that are consulted for $C_1$ (covering the period between 1800--1899) and $C_2$ (covering the period between 1946--1990), indicating the rise of a novel sense, the archaization of older senses or a change in frequency.
\end{itemize}
Three dictionaries are consulted throughout the analysis: (i) the Dictionary of the German language \citep{DWB} by Jacob und Wilhelm Grimm (digitized version of the 1st print published between 1854--1961), (ii) the Dictionary of Contemporary German \citep{WGD}, published between 1964--1977, now curated and digitized by the \citet{DWDS} and (iii) the Duden online dictionary of German language \citep{DUD}, reflecting usage of Contemporary German up until today.\footnote{Only the fully-digitized version of the \citet{DWB}'s first print was consulted for this evaluation, since a revised version has not been completed yet and is only available for lemmas starting with letters a--f.} Additionally, lemma entries in the Wiktionary online dictionary \citep{Wiktionary} are consulted to verify genuinely novel senses described in Section 8.1.

\subsection{Records of novel senses}
In the case of 17 target words, all senses identified by the system are included in at least one of the three dictionaries consulted for the analysis. In the four remaining cases, at least one novel sense of a word is neither paraphrased nor given as an example of semantically related senses in the dictionaries:

\paragraph{einbinden}
Reference to the integration or embedding of details on a topic, event, person in respect to a chronological order within written text or visual presentation (e.g. for an exhibition on an author) is judged as a novel sense in close semantic proximity to the old sense `to bind sth. into sth.', e.g. flowers into a bundle of flowers. \textit{einbinden} is also used in technical contexts, meaning `to (physically) implement parts of a construction or machine into their intended slots'.
\paragraph{niederschlagen} In cases where the verb \textit{niederschlagen} co-occurs with the verb particle \textit{auf} and the noun \textit{Flügel}, the verb refers to a bird's action of repeatedly moving its wings up and down in order to fly.
\paragraph{regelrecht} Used as an adverb, \textit{regelrecht} may refer to something being the usual outcome that ought to be expected due to scientific principles, with an emphasis on the actual result of an action (such as the dyeing of fiber of a piece of clothing following the bleaching process), whereas senses included in dictionaries for general language emphasize either the intended accordance with a rule or something usually happening (the latter being colloquial use). 
\paragraph{Zehner} (see Figure \ref{fig:zehner} in Appendix \ref{sec:appendix}) The meaning `a winning sequence of numbers in the national lottery', predicted to have risen as a novel sense between $C_1$ and $C_2$, is not included in any of the reference dictionaries.\\ 

\noindent In most of these cases, senses identified as novel reflect metaphoric use, indicating that definitions in existing dictionary entries may need to be broadened, or example sentences would have to be added. Some of the senses described in this section might be included in specialized dictionaries, e.g. technical usage of \textit{einbinden}.

\subsection{Records of changes} 
For 12 target words, semantic change predicted by the models (innovative, reductive or a salient change of frequency of a sense) correlates with the addition or non-inclusion of senses in dictionary entries consulted for the respective period of time (\citet{DWB} for $C_1$, \citet{WGD} for $C_2$). It should be noted though, that lemma lists of the two dictionaries might be lacking lemmas in the headword list, and lemma entries might be lacking paraphrases or examples of senses of the lemma, simply because corpus-based lexicography was not available at the time of their first print and revisions of the dictionaries are currently work in progress.

Additionally, we consult a dictionary for Early New High German \citep{FHDW} in order to check whether discovered novel senses existed at an earlier stage and may be discovered due to low frequency or sampling error.
In two cases, discovered novel senses that are not included in the \citet{DWB} (for $C_1$) are found to be included in the \citet{FHDW}. 

Interestingly, one sense paraphrased for \textit{Ausrufung} (`a loud wording, a shout') is included in neither of the two dictionaries consulted to judge senses from $C_1$ and $C_2$, but in the \citet{FHDW} (earlier) and \citet{DUD} (as of now). These findings suggest that it might be reasonable to use more than two reference corpora. This would also alleviate the corpus bias stemming from idiosyncratic data sampling procedures.

\section{Conclusion}
We used two state-of-the-art approaches to LSC detection in combination with a recently published high-quality data set to automatically discover semantic changes in a German diachronic corpus pair. While both approaches were able to discover various semantic changes with above-random probability, some of them previously undescribed in etymological dictionaries, the type-based approach showed a clearly better performance.

We validated model predictions by an optimized human annotation process yielding high inter-annotator agreement and providing convenient ways of visualization. In addition, we evaluated the full discovery process from a lexicographer's point of view and conclude that we obtained high-quality predictions, useful visualizations and previously unreported changes.
On the other hand, we discovered some issues with respect to the reliability of predictions for semantic change and number and composition of reference corpora that are going to be dealt with in the future. The results of the analyses endorse that our approach might aid lexicographers with extending and altering existing dictionary entries.

\section*{Acknowledgments}
We thank the three reviewers for their insightful feedback and Pedro González Bascoy for setting up the DURel annotation tool. Dominik Schlechtweg was supported by the Konrad Adenauer Foundation and the CRETA center funded by the German Ministry for Education and Research (BMBF) during the conduct of this study.

\bibliographystyle{acl_natbib}
\bibliography{Bibliography-general,bibliography-self,additional-references}

\clearpage
\appendix

\section{Appendix}
\label{sec:appendix}

\begin{figure*}
    \begin{subfigure}{0.33\textwidth}
\frame {        \includegraphics[width=\linewidth]{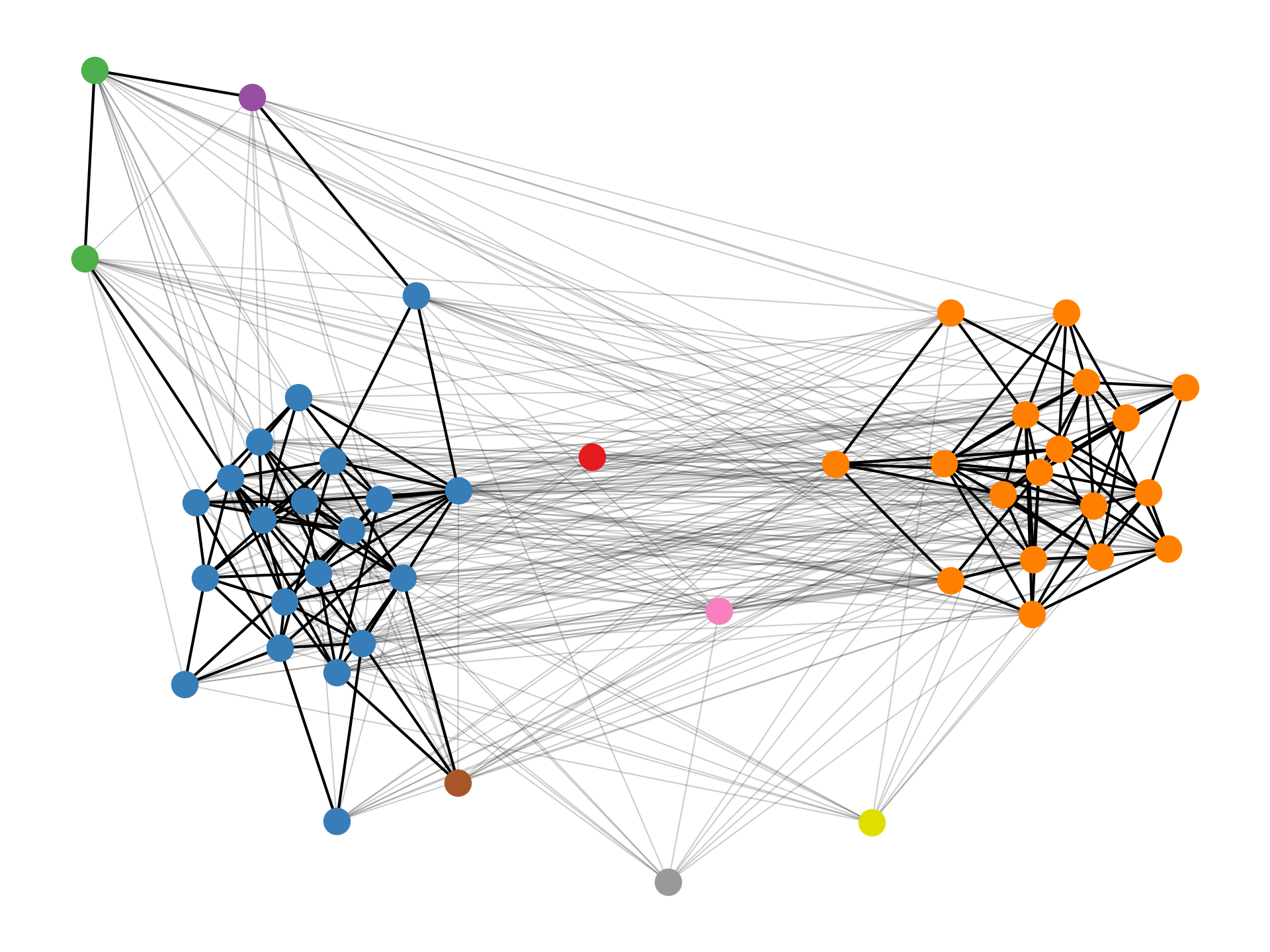}}
        \caption*{full}
    \end{subfigure}
    \begin{subfigure}{0.33\textwidth}
\frame{        \includegraphics[width=\linewidth]{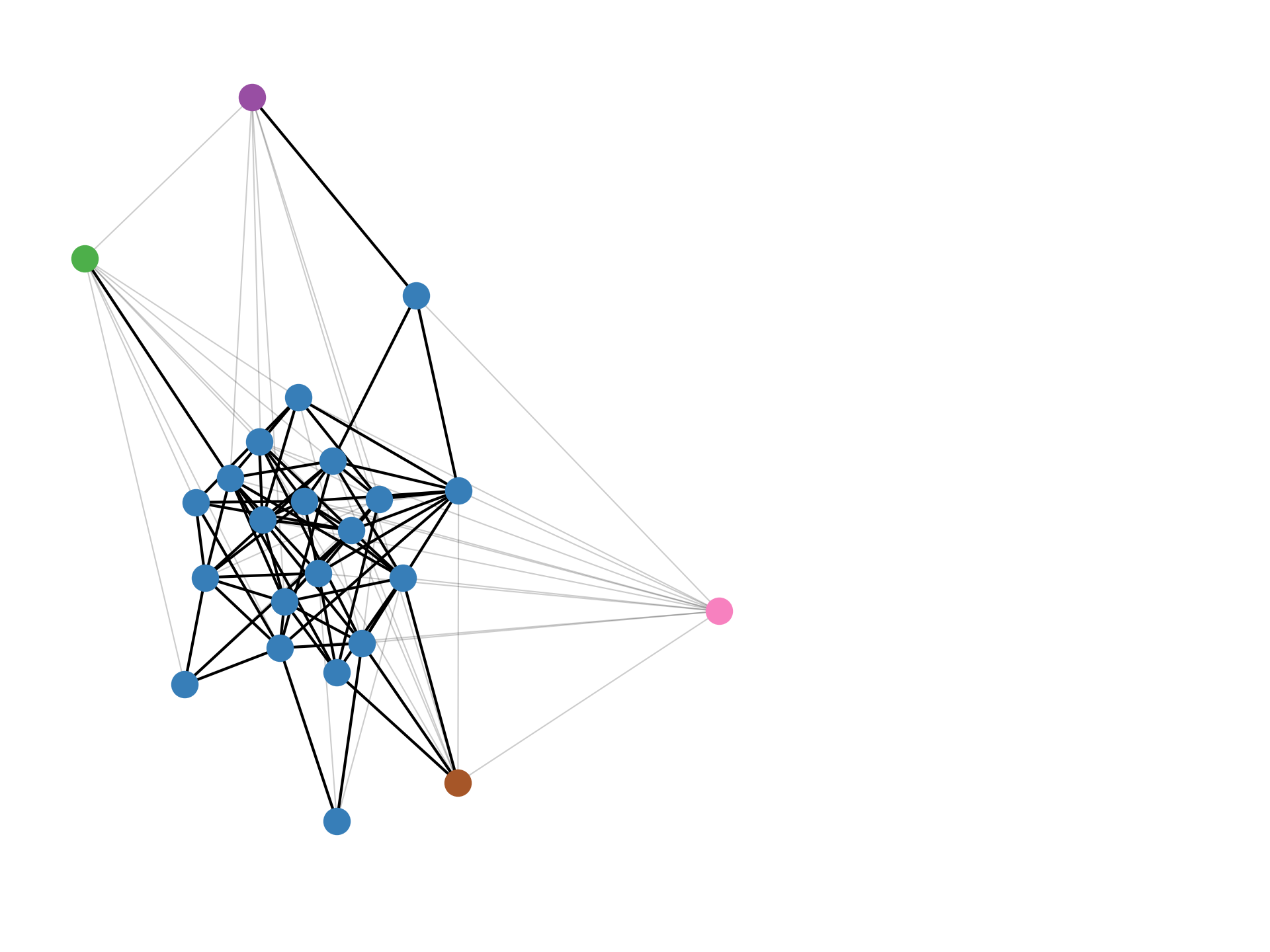}}
        \caption*{$C_1$}
    \end{subfigure}
    \begin{subfigure}{0.33\textwidth}
\frame{        \includegraphics[width=\linewidth]{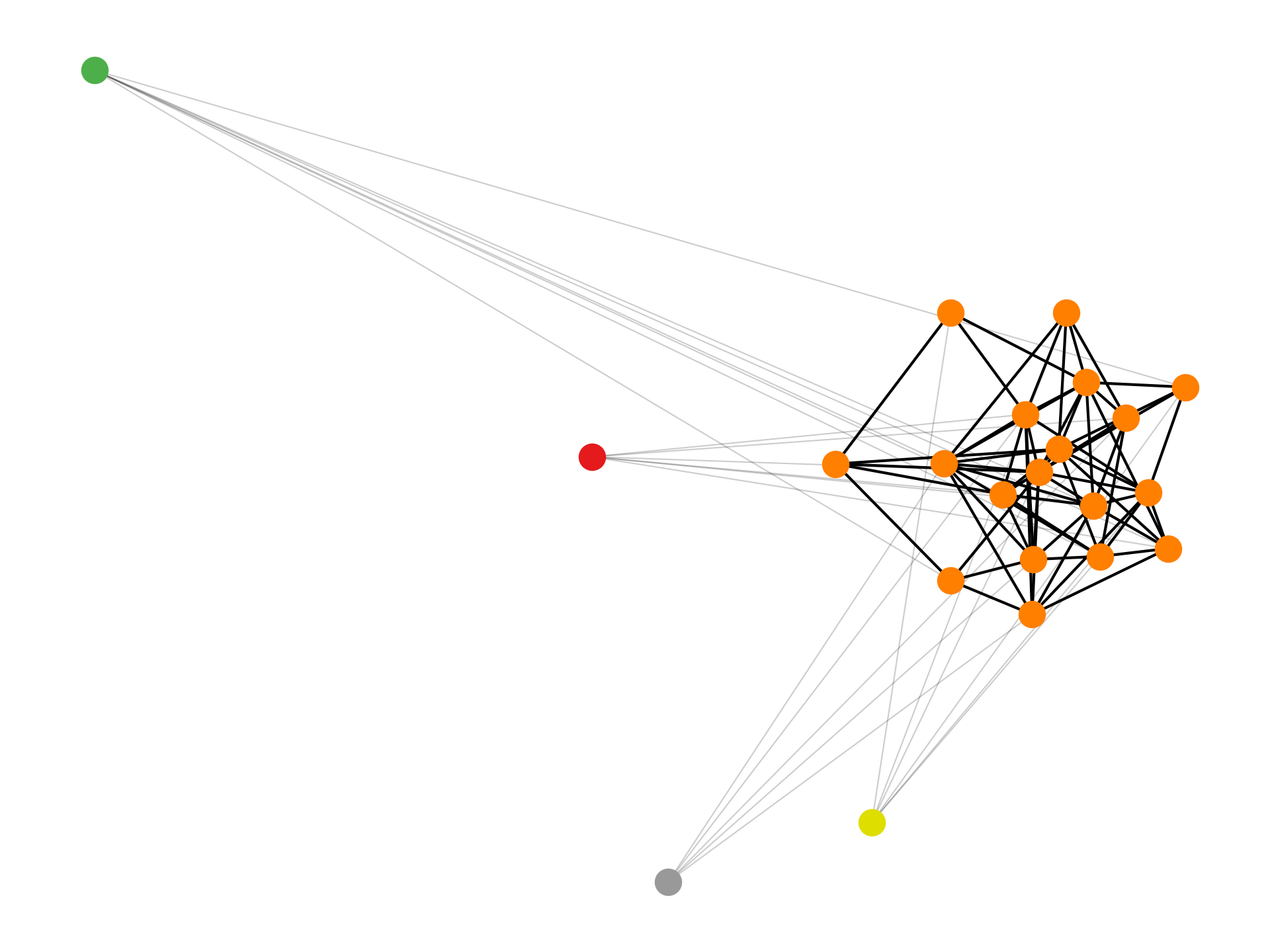}}
        \caption*{$C_2$}
    \end{subfigure}
    \vspace{-2mm}
    \caption{Word Usage Graph of German \textit{Zehner} (left), subgraphs for first time period $C_1$ (middle) and for second time period $C_2$ (right).}\label{fig:zehner}
\end{figure*}

\begin{figure*}
    \begin{subfigure}{0.33\textwidth}
\frame {        \includegraphics[width=\linewidth]{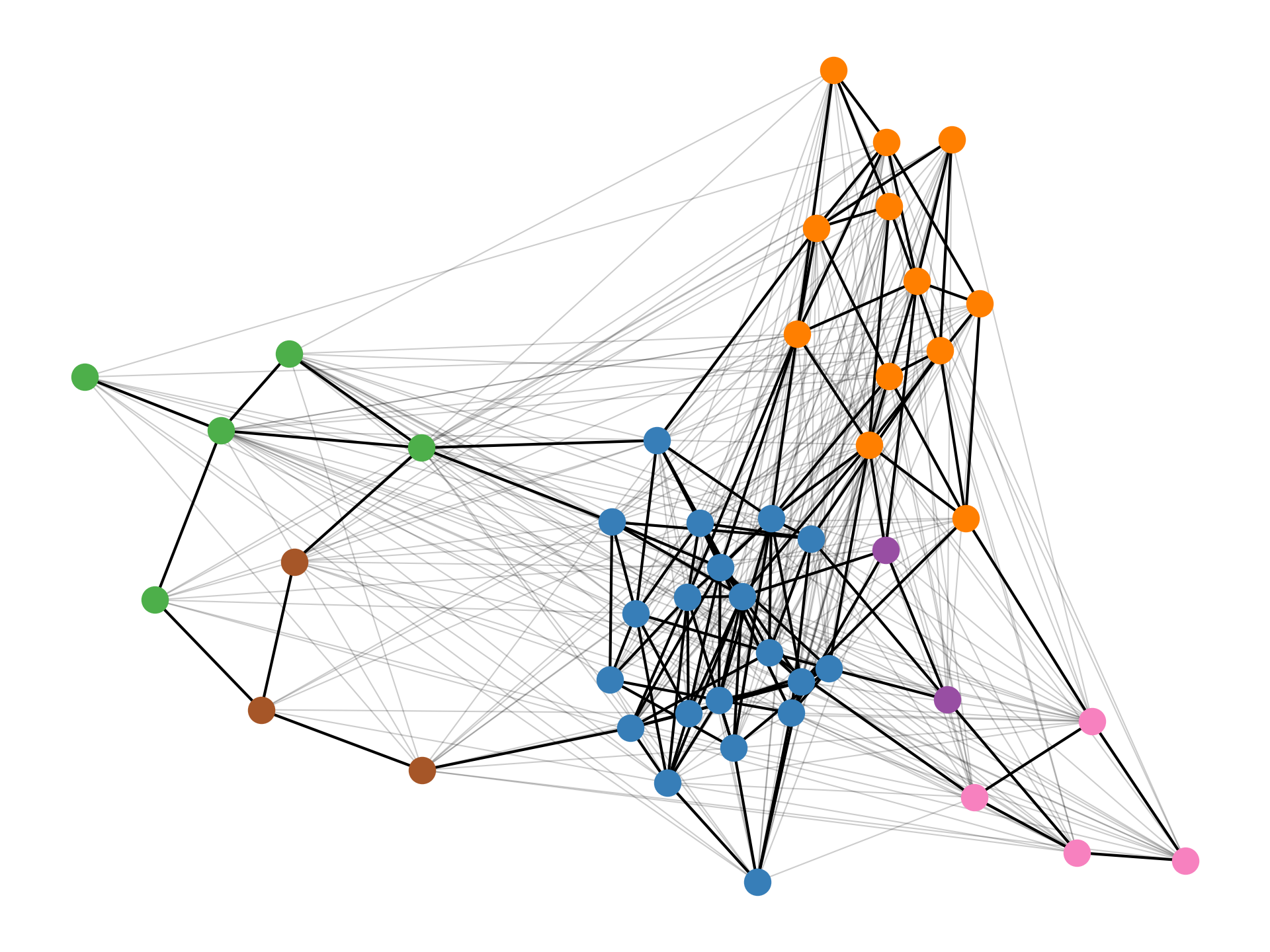}}
        \caption*{full}
    \end{subfigure}
    \begin{subfigure}{0.33\textwidth}
\frame{        \includegraphics[width=\linewidth]{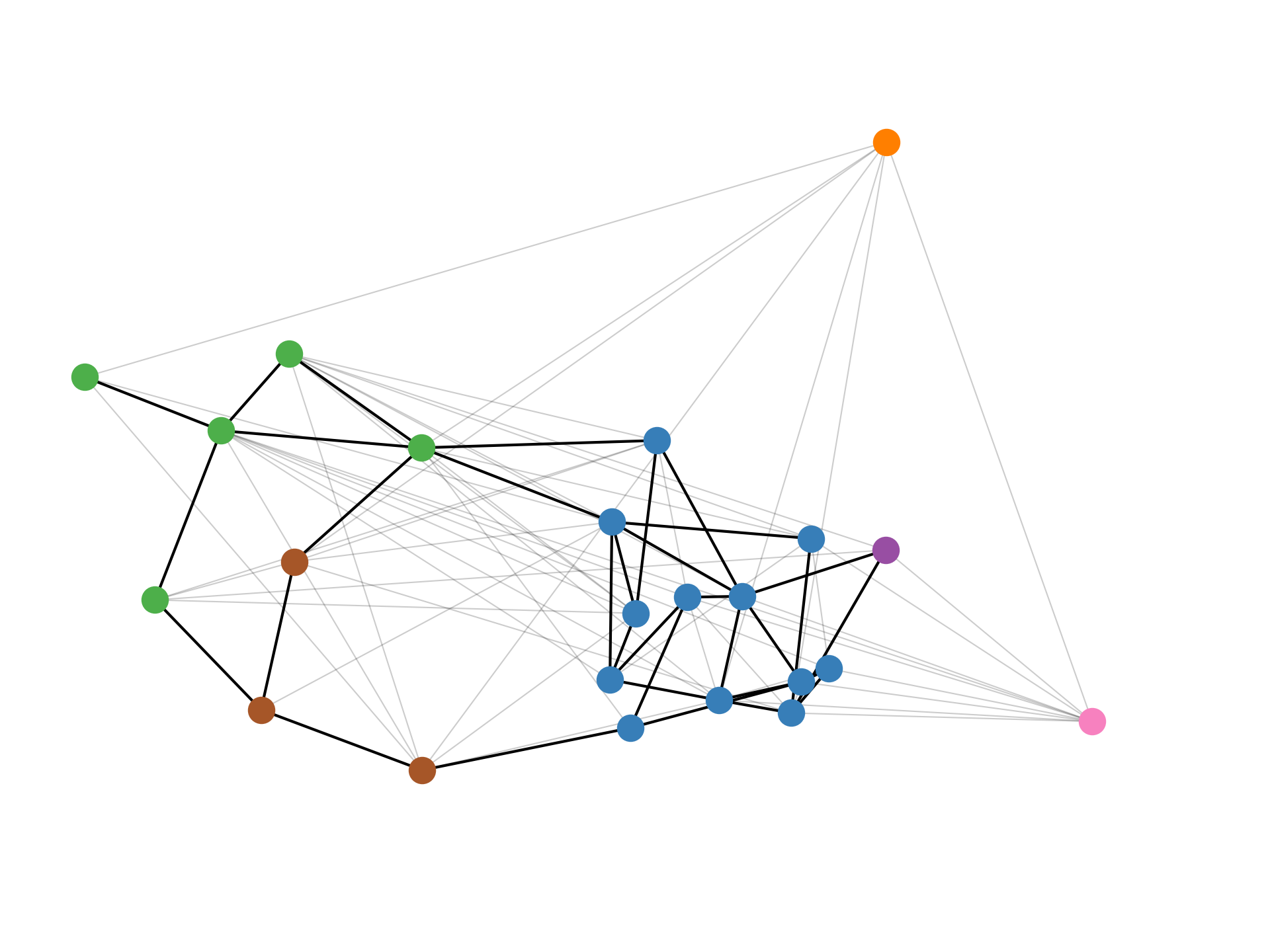}}
        \caption*{$C_1$}
    \end{subfigure}
    \begin{subfigure}{0.33\textwidth}
\frame{        \includegraphics[width=\linewidth]{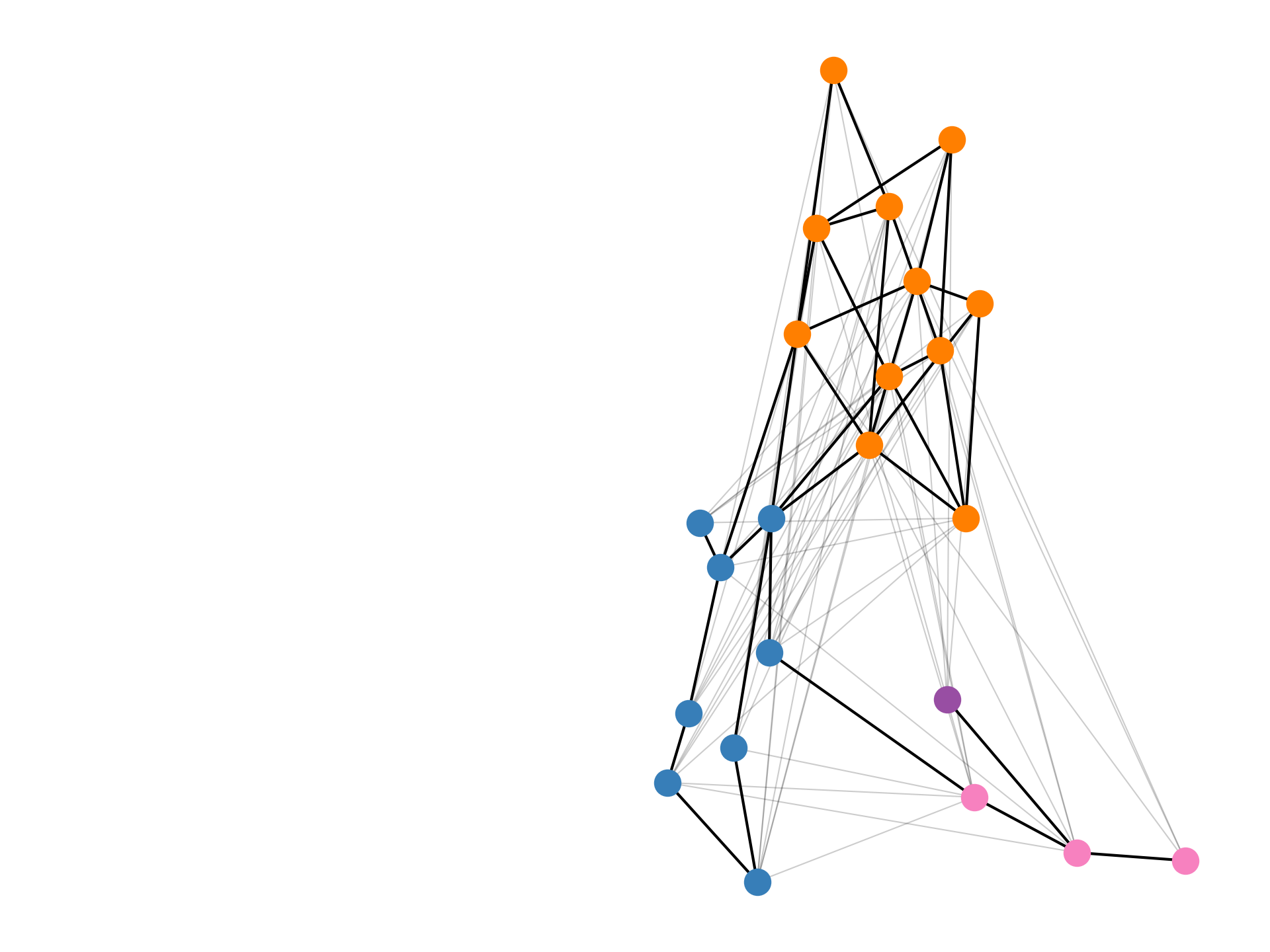}}
        \caption*{$C_2$}
    \end{subfigure}
    \vspace{-2mm}
    \caption{Word Usage Graph of German \textit{Lager} (left), subgraphs for first time period $C_1$ (middle) and for second time period $C_2$ (right).}\label{fig:lager}
\end{figure*}

\begin{figure*}
    \begin{subfigure}{0.33\textwidth}
\frame {        \includegraphics[width=\linewidth]{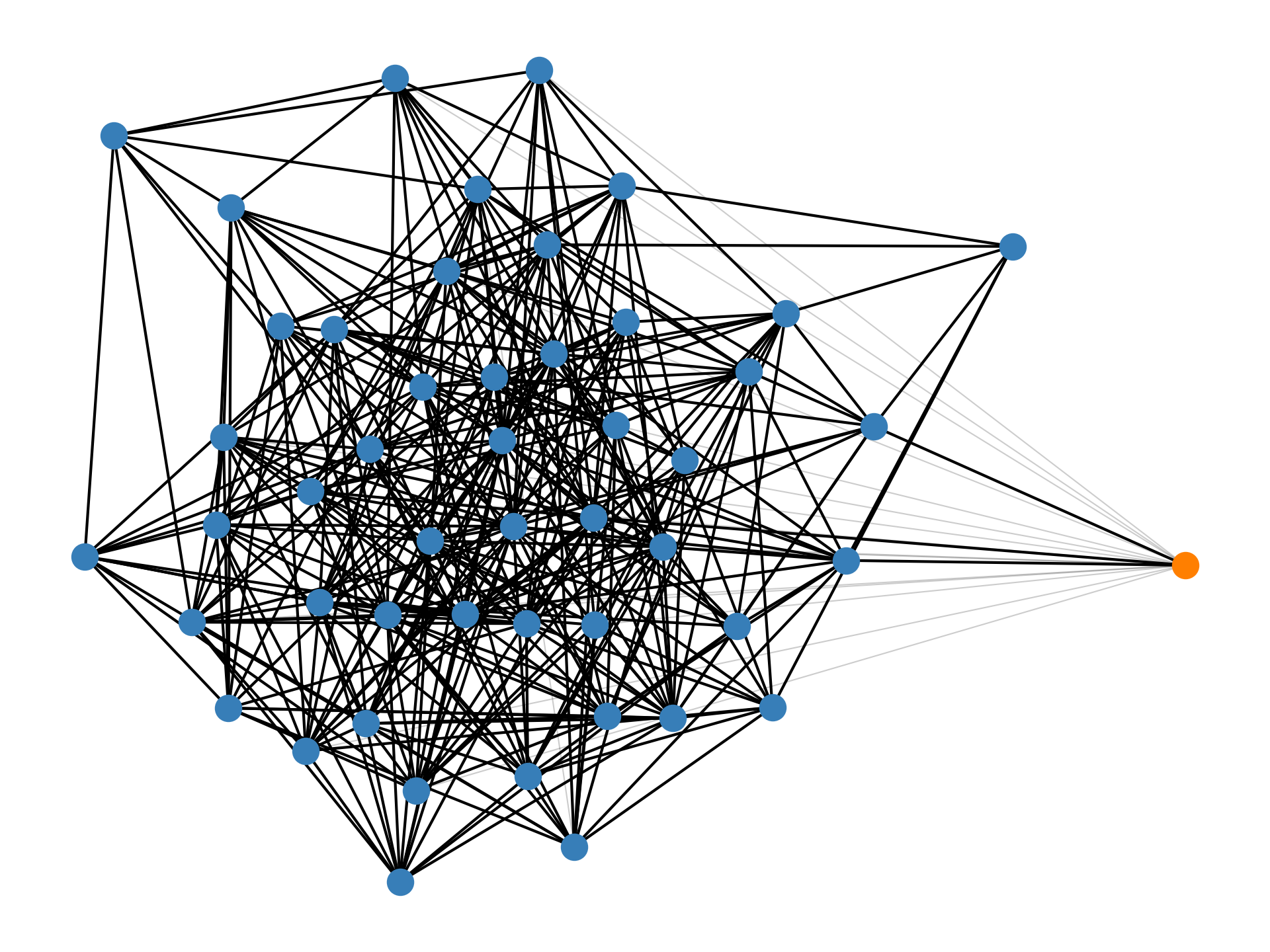}}
        \caption*{full}
    \end{subfigure}
    \begin{subfigure}{0.33\textwidth}
\frame{        \includegraphics[width=\linewidth]{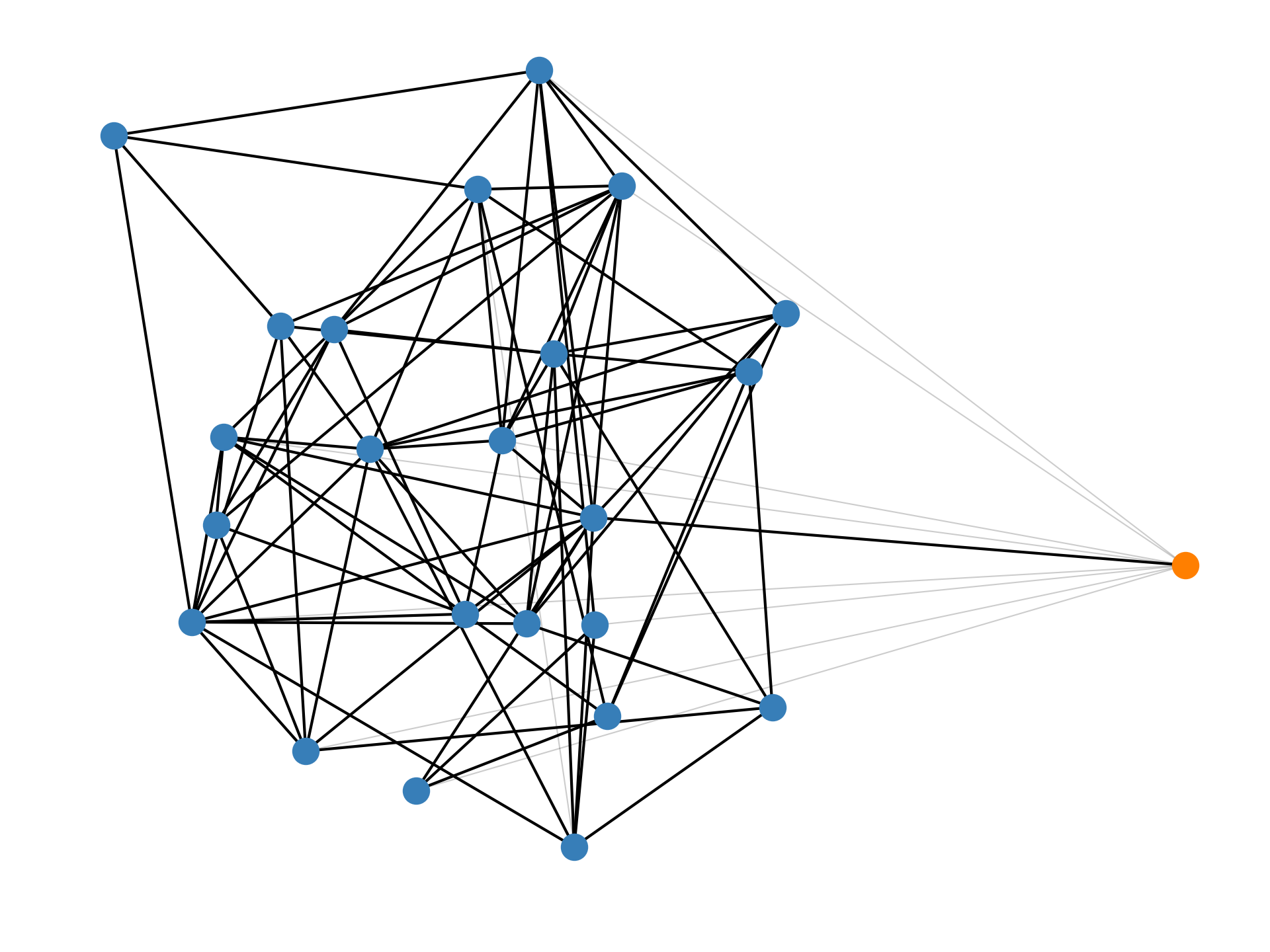}}
        \caption*{$C_1$}
    \end{subfigure}
    \begin{subfigure}{0.33\textwidth}
\frame{        \includegraphics[width=\linewidth]{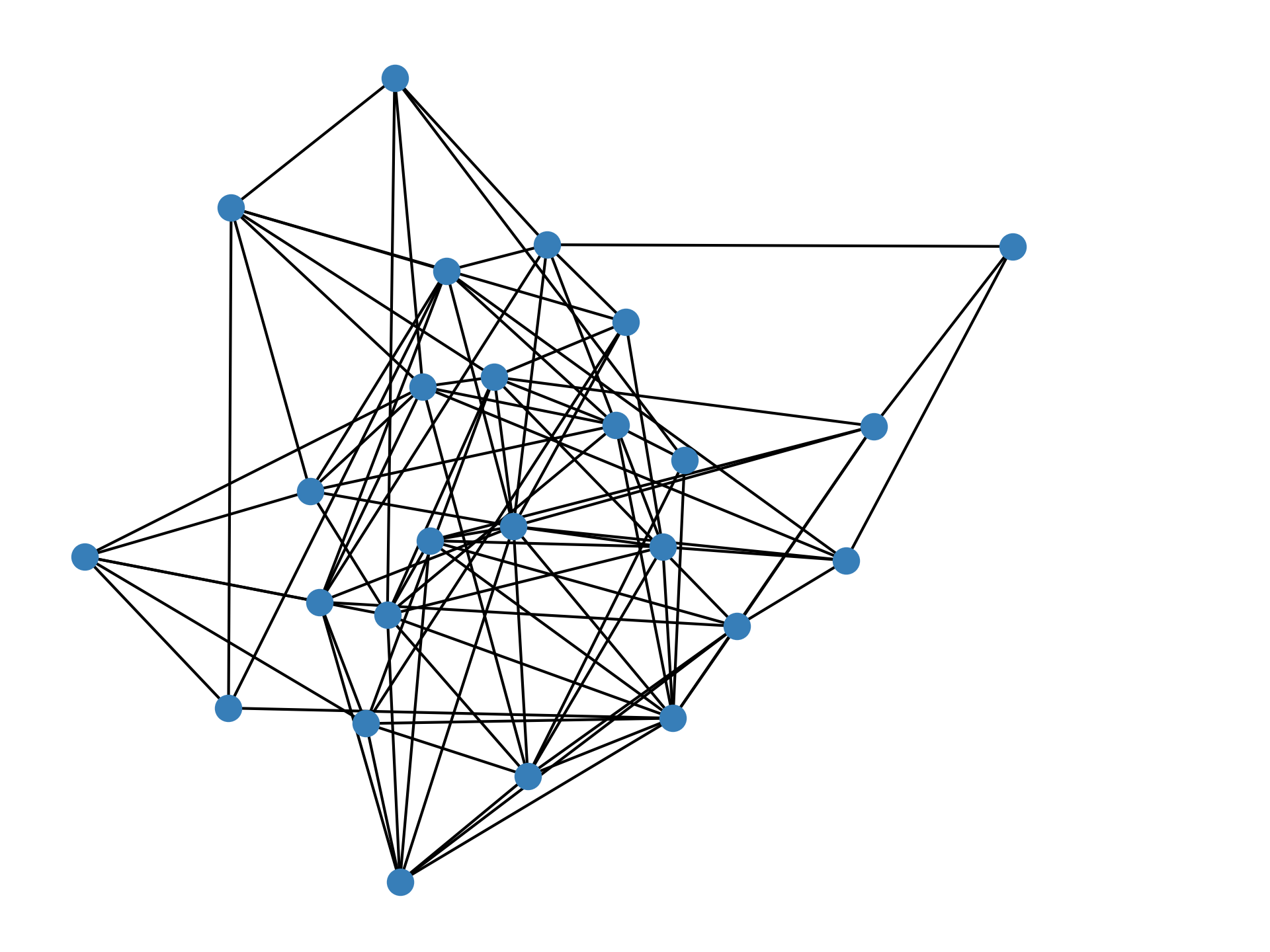}}
        \caption*{$C_2$}
    \end{subfigure}
    \vspace{-2mm}
    \caption{Word Usage Graph of German \textit{Anriffswaffe} (left), subgraphs for first time period $C_1$ (middle) and for second time period $C_2$ (right).}\label{fig:angriffswaffe}
\end{figure*}

\begin{figure*}
    \begin{subfigure}{0.33\textwidth}
\frame {        \includegraphics[width=\linewidth]{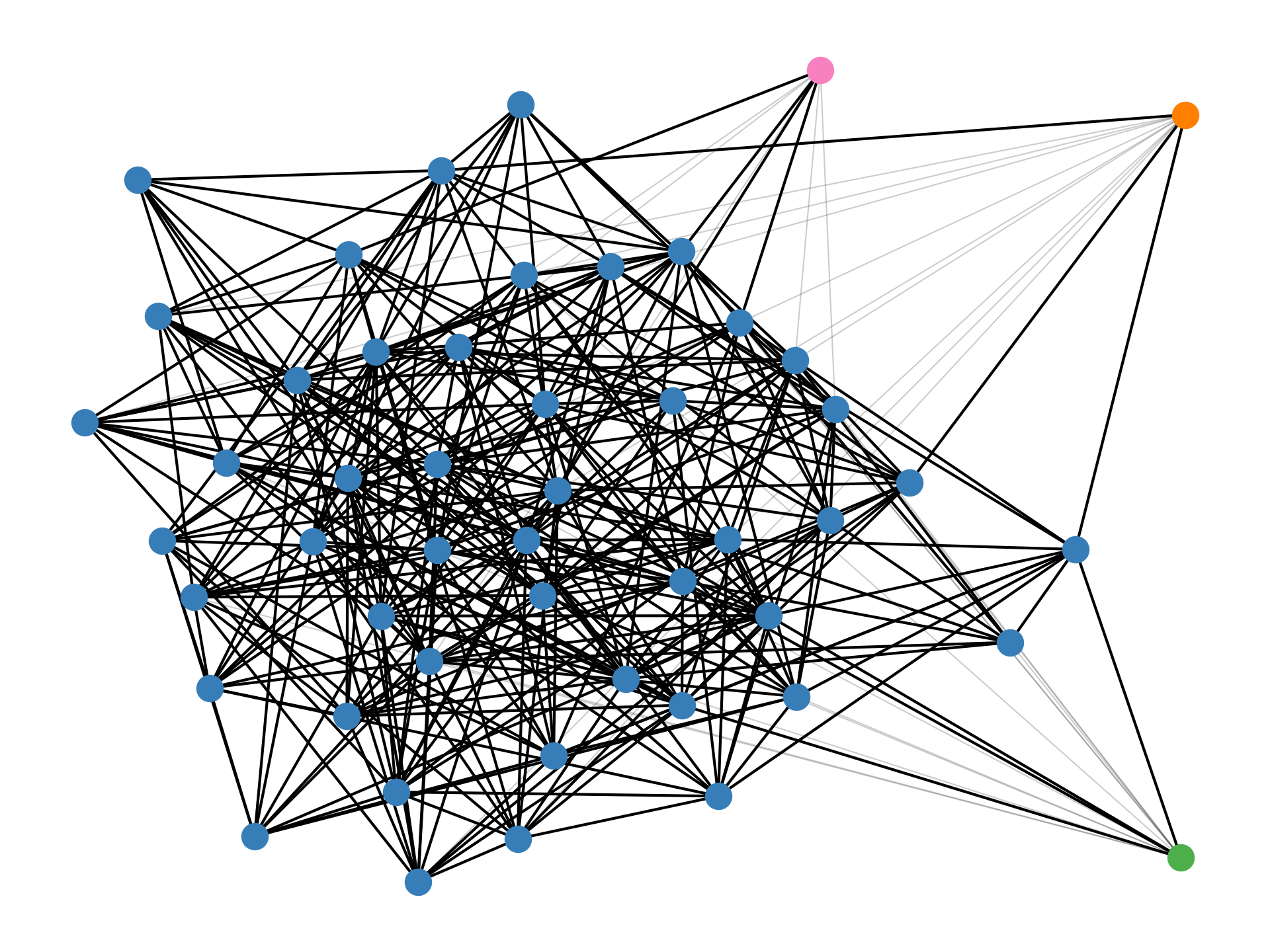}}
        \caption*{full}
    \end{subfigure}
    \begin{subfigure}{0.33\textwidth}
\frame{        \includegraphics[width=\linewidth]{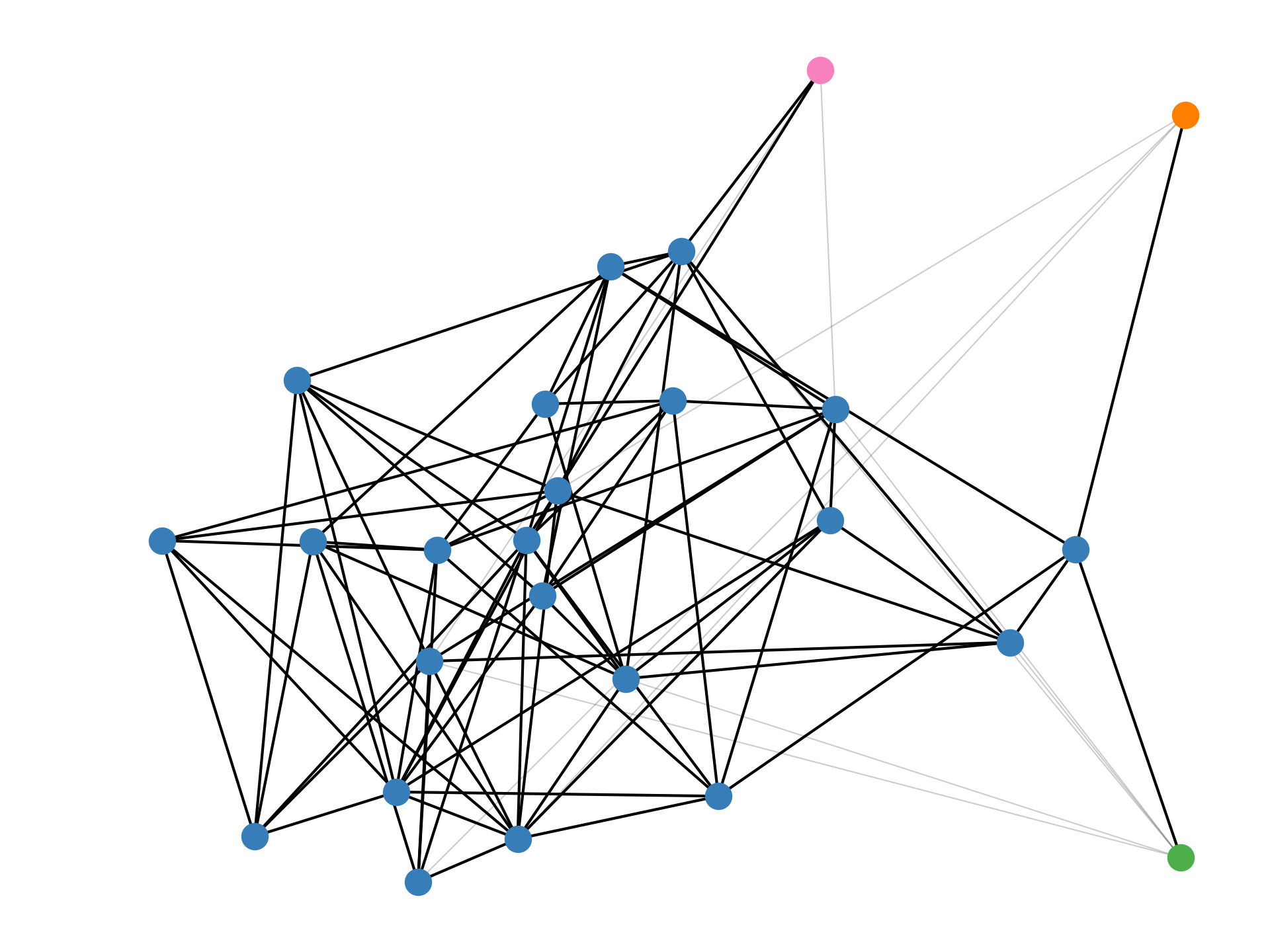}}
        \caption*{$C_1$}
    \end{subfigure}
    \begin{subfigure}{0.33\textwidth}
\frame{        \includegraphics[width=\linewidth]{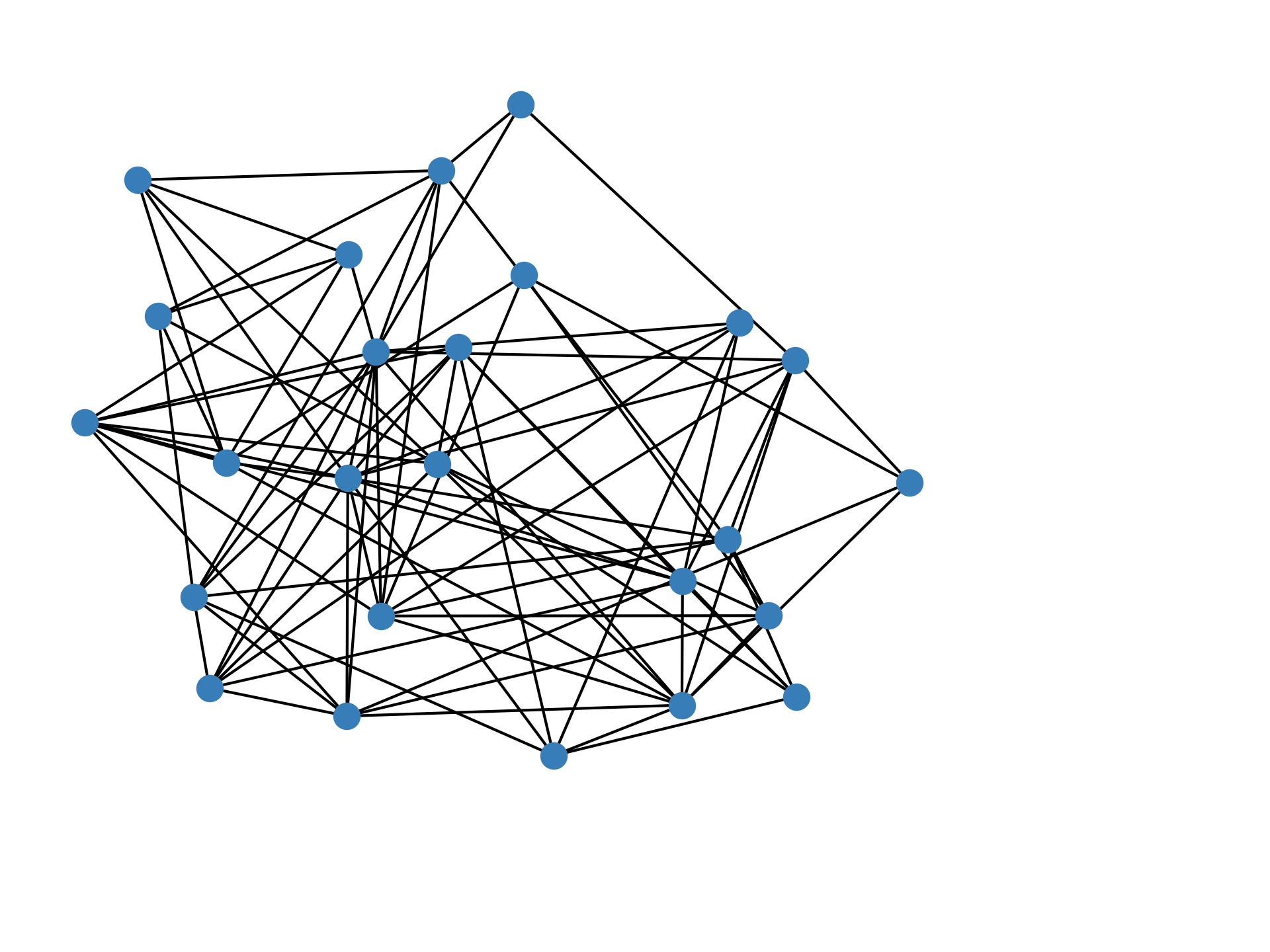}}
        \caption*{$C_2$}
    \end{subfigure}
    \vspace{-2mm}
    \caption{Word Usage Graph of German \textit{neunjährig} (left), subgraphs for first time period $C_1$ (middle) and for second time period $C_2$ (right).}\label{fig:neunjahrig}
\end{figure*}

\end{document}